\documentclass{article}

\usepackage{microtype}
\usepackage{graphicx}
\usepackage{subfigure}
\usepackage{booktabs} 
\usepackage{hyperref}

\usepackage{booktabs} 
\usepackage{caption}  
\usepackage{multirow}
\usepackage{array}
\usepackage{svg}       
\usepackage{graphicx}  
\usepackage{enumitem}
\usepackage{svg}
\usepackage{fontawesome5}
\usepackage[table]{xcolor} 

\definecolor{mygray}{gray}{0.9}
\usepackage{tcolorbox}
\usepackage{xcolor}

\usepackage[accepted]{icml2025}

\usepackage{amsmath}
\usepackage{amssymb}
\usepackage{mathtools}
\usepackage{amsthm}

\usepackage[capitalize,noabbrev]{cleveref}

\theoremstyle{plain}

\theoremstyle{definition}

\theoremstyle{remark}

\usepackage[textsize=tiny]{todonotes}

\icmltitlerunning{Non-Markovian Long-Horizon Robot Manipulation via Keyframe Chaining}

\begin{document}

\twocolumn[
\icmltitle{Non-Markovian Long-Horizon Robot Manipulation via Keyframe Chaining}



\icmlsetsymbol{equal}{*}
\icmlsetsymbol{corr}{\faEnvelope}

\begin{icmlauthorlist}
\icmlauthor{Yipeng Chen}{tongji,equal}
\icmlauthor{Wentao Tan}{tongji,equal}
\icmlauthor{Lei Zhu}{tongji,corr}
\icmlauthor{Fengling Li}{uts}
\icmlauthor{Jingjing Li}{uestc}
\icmlauthor{Guoli Yang}{aibd}
\icmlauthor{Heng Tao Shen}{tongji}
\end{icmlauthorlist}

\icmlaffiliation{tongji}{Tongji University}
\icmlaffiliation{uts}{University of Technology Sydney}
\icmlaffiliation{uestc}{University of Electronic Science and Technology of China}
\icmlaffiliation{aibd}{Advanced Institute of Big Data}

\icmlcorrespondingauthor{Lei Zhu}{leizhu0608@gmail.com}

\icmlkeywords{Machine Learning, Robot Learning}

\vskip 0.3in
]

\printAffiliationsAndNotice{\icmlEqualContribution}



\begin{abstract}
Existing Vision-Language-Action (VLA) models often struggle to generalize to long-horizon tasks due to their heavy reliance on immediate observations. While recent studies incorporate retrieval mechanisms or extend context windows to handle procedural tasks, they often struggle to capture Non-Markovian dependencies, where optimal actions rely solely on specific past states rather than the current observation. To address this, we introduce Keyframe-Chaining VLA, a framework that extracts and links key historical frames to model long-horizon dependencies. Specifically, we propose an automatic keyframe selector that learns a discriminative embedding space, effectively identifying distinct state transitions. To capture task-critical information, we design a progress-aware query mechanism that dynamically retrieves historical frames based on their temporal relevance to the current execution phase. These selected keyframes are integrated into the VLA as interleaved visual tokens, explicitly grounding the policy in the long-horizon temporal context. Finally, we introduce a suite of four Non-Markovian manipulation tasks built upon the ManiSkill simulator to measure task success rates. Experimental results demonstrate that our method achieves superior performance, effectively tackling robot manipulation tasks characterized by long-horizon temporal dependencies. Code is available at \url{https://github.com/cytoplastm/KC-VLA}.
\end{abstract}

\begin{figure*}[t]
    \centering
    \includegraphics[width=0.95\textwidth]{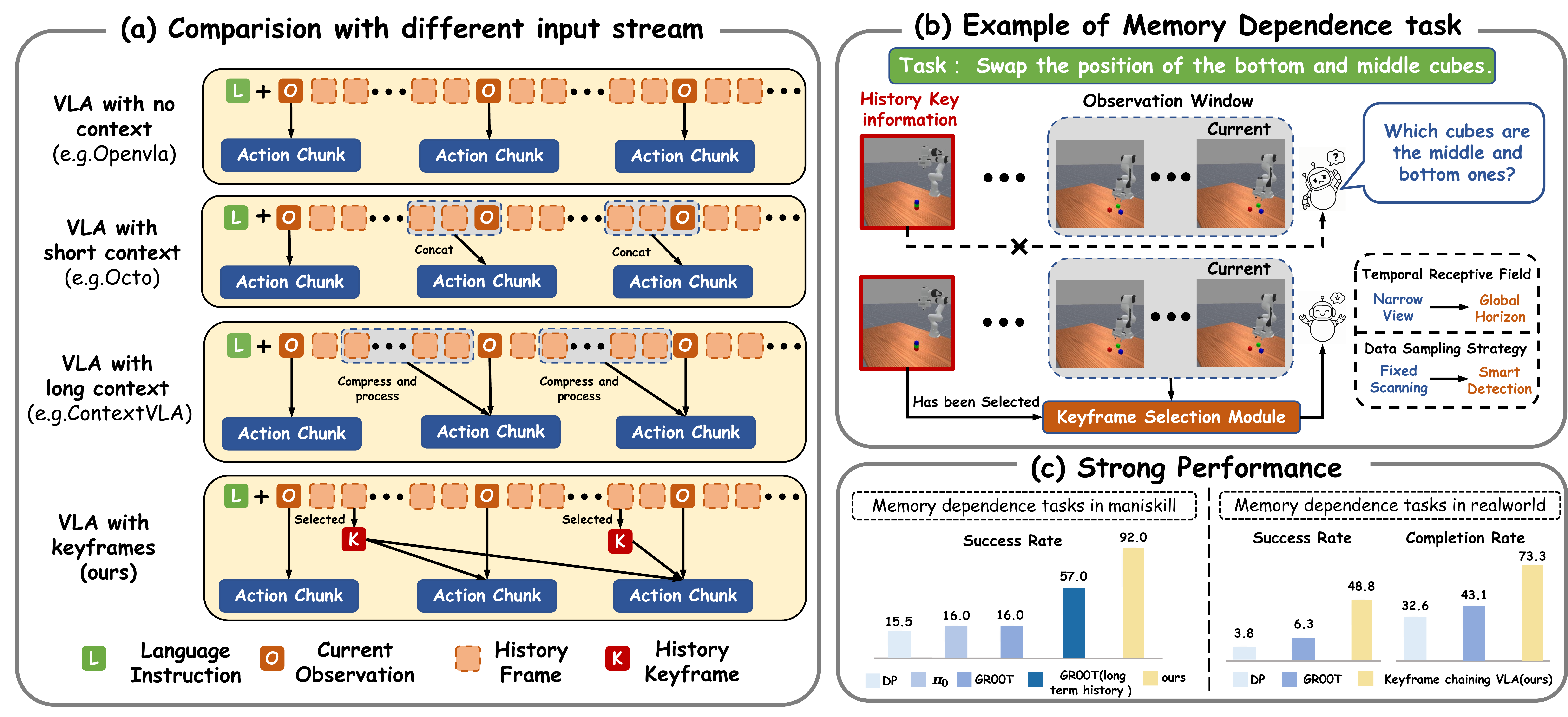}
\caption{
\textbf{(a)} Unlike standard VLAs ~\cite{openvla,octo,dp} that rely on dense sliding windows or context compression~\cite{contextvla,hamlet}, our Keyframe-Chaining VLA explicitly selects sparse semantic keyframes, avoiding computational redundancy.
\textbf{(b)} The other three paradigms in (a) suffer from limited temporal receptive fields, failing to capture distant critical cues. In contrast, our KSM-based retrieval achieves a global temporal receptive field, enabling the agent to access long-range historical information and effectively resolve state aliasing.
\textbf{(c)} Our framework achieves state-of-the-art performance, significantly outperforming baselines in both ManiSkill memory-dependent tasks and real-world long-horizon deployment.
}
    \label{fig:workflow}
\end{figure*}

\section{Introduction}
State-of-the-art Vision-Language-Action (VLA)~\cite{gr00t,pi0,octo} models have demonstrated remarkable control capabilities but remain shackled by a ``dense observation" paradigm. Relying on short, dense history windows implicitly assumes a \textit{Markovian property}, which collapses in long-horizon tasks characterized by state aliasing—where the immediate observation fails to reflect critical past interactions~\cite{pomdp}. While extending the context window theoretically mitigates this, it hits the computational wall of quadratic attention complexity, rendering real-time control prohibitive. Existing solutions, such as latent compression or hierarchical VLM planning, often compromise either fine-grained spatial precision or inference speed.

To transcend these limitations, we propose the \textbf{Keyframe-Chaining VLA}, a framework that decouples temporal abstraction from action generation. Instead of burdening the policy with redundant video streams, we introduce a lightweight \textbf{Keyframe Selection Module (KSM)} to distill continuous observations into a Sparse Semantic History composed of critical semantic keyframes. This allows the policy to reason over events separated by thousands of timesteps as effortlessly as adjacent frames, effectively `flattening' the temporal horizon to establish a global perspective of the task structure. 

The KSM features a novel Task-Modulated architecture: by leveraging Feature-wise Linear Modulation (FiLM) within a unified metric space, it dynamically adjusts retrieval criteria based on task semantics to filter motion noise. This sparse, semantically aligned context is then injected into a flow-matching policy via a context-aware prompting strategy. This paradigm enables the agent to reason over global task topology with negligible computational overhead.

Our contributions are summarized as follows: We propose the Keyframe-Chaining VLA, a framework that resolves non-Markovian ambiguity via sparse semantic history, bypassing the compute bottleneck of long-context attention. We design a Task-Modulated KSM utilizing FiLM-based query generation to extract semantic keyframes in a unified multi-task metric space.  We establish a specialized ManiSkill Benchmark Suite tailored for Long-horizon Semantic Reasoning, bridging the gap in existing benchmarks which predominantly focus on Markovian settings. Extensive experiments demonstrate that our method achieves a 92\% success rate on these memory-intensive tasks, significantly outperforming  baselines (57\%). 

\section{Related Works}

\subsection{Vision-Language-Action Models}
Pioneering VLAs such as RT-2~\cite{rt2} and OpenVLA~\cite{openvla} employ discrete tokenization to achieve semantic generalization, while recent state-of-the-art models like $\pi_0$~\cite{pi0} and GR00T~\cite{gr00t} shift towards continuous generation via diffusion or flow matching to enhance action smoothness. Despite these architectural evolutions, the dominant input paradigm remains constrained to a singleton current frame or a dense, short-term history window. This design, necessitated by the quadratic cost of attention mechanisms, inherently restricts the temporal receptive field, rendering even advanced foundational agents ``myopic" and ill-equipped for tasks requiring extensive long-horizon reasoning.

\subsection{Memory Mechanisms for Long-Context Policies}
To incorporate historical context, recent research explores three primary strategies. First, token compression methods like ContextVLA~\cite{contextvla} and HAMLET~\cite{hamlet} map observations into compact clusters. However, their compression ratios are often insufficient to cover full-episode histories, leading to the loss of early cues. Second, hierarchical planners such as MemER~\cite{memer} utilize VLMs to generate textual subgoals. While logically robust, this reliance on natural language introduces a bottleneck for fine-grained spatial information and incurs prohibitive inference overhead. Third, explicit visual retention methods focus on streaming memory or tracking. MemoryVLA~\cite{memoryvla} constructs a memory bank but suffers from information attenuation due to recursive fusion. Similarly, TraceVLA~\cite{tracevla}, SAM2Act~\cite{sam2act}, and HiF-VLA~\cite{hifvla} explicitly track physical dynamics but lack the capacity to abstract high-level task semantics. Distinct from these, our framework employs a lightweight selection module to explicitly retain a discrete sequence of semantic keyframes spanning the entire episode, avoiding both information blurring and computational redundancy.

\subsection{Video Keyframe Selection}
The field of computer vision has extensively explored long-context video understanding~\cite{videollama,moviechat}, with recent works such as AKS~\cite{adaptive} and frame-voyager~\cite{framevoyager} employing keyframe selection to enhance efficiency. However, these approaches typically rely on heavy multimodal encoders or iterative VLM queries, incurring high per-frame computational costs and inference latency. Such characteristics are ill-suited for closed-loop robotic control, which imposes strict real-time latency constraints. Furthermore, traditional video summarization algorithms like DSNet~\cite{dsnet}, AC-SUM-GAN~\cite{acsumgan}, and DR-DSN~\cite{drdsn} primarily operate in offline settings, relying on global optimization over the full sequence. This fundamentally conflicts with the streaming nature of robotic manipulation. Designed explicitly for real-time control, our method employs a lightweight metric learning module to achieve online, streaming extraction of semantic keyframes with negligible computational overhead.

\section{Preliminaries}

\subsection{Language-Conditioned Control Policies}

Language-conditioned robotic policies are typically trained to model the conditional distribution $\pi(\mathbf{A}_t \mid o_t)$, where $\mathbf{A}_t = [a_t, a_{t+1}, \dots, a_{t+H-1}]$ denotes an action chunk spanning $H$ future time steps starting from the current time $t$. The variable $o_t$ represents the robot's current sensory observation. Specifically, the observation is formulated as $o_t = [\mathbf{I}_t, l, q_t]$, where $\mathbf{I}_t = [I_t^1, I_t^2, \dots, I_t^N]$ comprises images from $N$ camera viewpoints, $l$ is the natural language instruction, and $q_t$ denotes proprioceptive inputs (e.g., joint angles and gripper states).

\subsection{Non-Markovian Robotic Manipulation}

In standard short-horizon tasks, the current observation $o_t$ often contains sufficient statistics to infer the underlying system state, satisfying the Markov property. In such scenarios, a policy modeled as $\pi(\mathbf{A}_t \mid o_t)$ is sufficient for successful execution. However, complex manipulation tasks often introduce \textbf{Non-Markovian} challenges characterized by partial observability.

For example, in a cooking task, determining whether salt has already been added requires recalling past events, as the visual state of the soup may remain unchanged. In these cases, the immediate observation $o_t$ fails to fully reflect the true environmental state (a phenomenon known as \textit{state aliasing}), rendering the instantaneous policy $\pi(\mathbf{A}_t \mid o_t)$ ineffective. To solve such tasks, the policy must condition on the interaction history, formulated as $\pi(\mathbf{A}_t \mid o_{0:t})$, to resolve state ambiguity using temporal context.
\subsection{VLA Input Modalities: From Dense to Sparse}

Vision-Language-Action (VLA) models serve as the backbone policy, aiming to map multimodal inputs to continuous robot actions. The input modality fundamentally determines the model's temporal reasoning capabilities.

\textbf{Traditional Input Stream:} Conventional VLAs typically utilize either the singleton current frame or a concatenation of a short history window, modeled as $\pi(\mathbf{A}_t \mid o_t)$ or $\pi(\mathbf{A}_t \mid o_{t-k:t})$. Due to the quadratic computational complexity of attention mechanisms, the observation window size $k$ is strictly limited. Consequently, these models often fail in Non-Markovian tasks as they inevitably discard critical events occurring outside the dense sliding window.

\textbf{Keyframe-Driven Input Stream (Ours):} To overcome the horizon limitation, we propose a sparse, semantic-aware input formulation. Instead of a dense queue, we maintain a memory buffer of \textit{keyframes}—discrete observations captured at critical timestamps. Our objective is to ensure that the combination of historical keyframes and the current observation captures all critical historical information, approximating the full history: $\{ o_{k_1}, o_{k_2}, \dots, o_{k_n}, o_t \} \approx \{ o_{0:t} \}$. Accordingly, the VLA input stream is reformulated as $\pi(\mathbf{A}_t \mid o_{k_1}, o_{k_2}, \dots, o_{k_n}, o_t)$. This paradigm enables the model to reason over long-horizon dependencies with efficient memory representation.

\section{Method}

\begin{figure*}[t]
    \centering
    \includegraphics[width=0.95\textwidth]{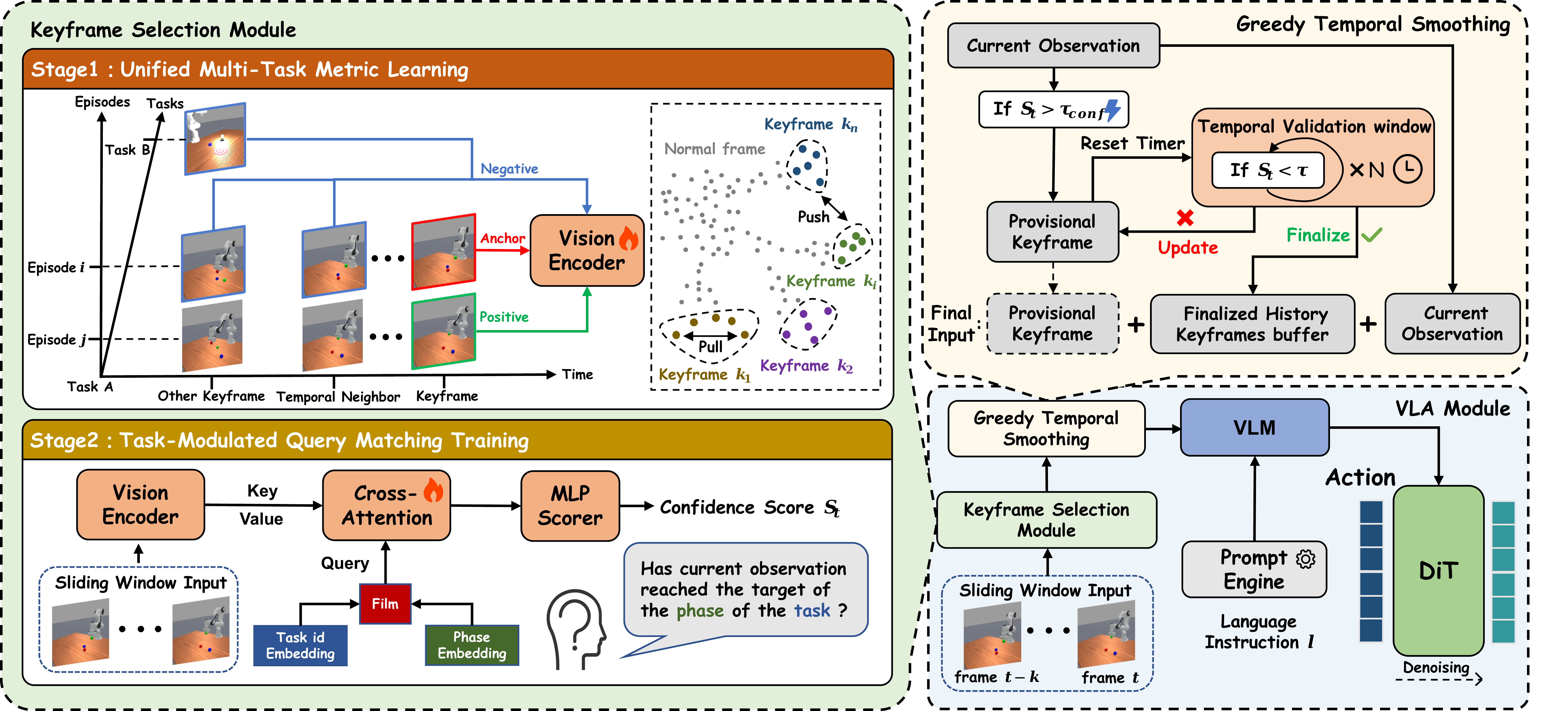}
\caption{
\textbf{Overview of the Keyframe-Chaining VLA framework.}
The architecture operates via two synergistic modules.
\textbf{(Left) Keyframe Selection Module:} Adopts a two-stage design where \textit{Stage I} optimizes a visual encoder via unified metric learning, and \textit{Stage II} employs a Task-Modulated Query network to compute phase-specific confidence scores $s_t$.
\textbf{(Top Right) Greedy Temporal Smoothing:} A stabilizing mechanism designed to filter detection jitter. It greedily updates provisional candidates to capture the latest valid state and finalizes the keyframe only after passing a temporal validation window.
\textbf{(Bottom Right) VLA Module:} The policy integrates the sparse semantic history of detected keyframes with the current observation to generate actions via Flow Matching.
}
    \label{fig:framework}
\end{figure*}

\subsection{System Overview}
We propose the Keyframe-Chaining VLA framework. As illustrated in Fig.~\ref{fig:framework}, the architecture materializes the sparse input paradigm through two synergistic modules: a lightweight, unified Keyframe Selection Module (KSM) and a Keyframe-Conditioned Action Policy. Specifically, the KSM acts as a temporal filter operating on the visual stream $\{\mathbf{I}_0, \dots, \mathbf{I}_t\}$. Upon detecting a semantic transition based on visual cues, the module explicitly indexes and retains the corresponding full multimodal observation $o_t$ (including vision and proprioception) to form the discrete sequence of semantic keyframes $\mathcal{O}_{\text{key}} = \{o_{k_1}, \dots, o_{k_n}\}$.Subsequently, the policy synthesizes actions conditioned on these retrieved keyframes and the current observation, utilizing a structured input format to organize the multimodal context. This decoupled design enables efficient long-horizon reasoning without altering the backbone's internal architecture.

\subsection{Keyframe Selection Module}

We propose a two-stage Task-Modulated architecture to extract semantic keyframes. Formally, we model each long-horizon task as a strictly ordered sequence of semantic phases. We define Keyframes as the canonical states marking the completion of each phase. Consequently, a task with $N$ phases is characterized by $N$ ground-truth keyframes. The detection of the $k$-th keyframe functionally serves as a transition trigger, advancing the system from phase $k$ to $k+1$. These keyframe annotations are acquired via a hybrid strategy: for tasks where semantic keyframes are accompanied by distinct changes in proprioceptive states, we leverage these signals to automate keyframe extraction. Otherwise, we employ manual annotation to ensure accuracy. This module learns a unified latent space, allowing a dynamic, task-conditioned query to identify these critical state transitions in real-time.

\noindent\textbf{Stage I: Unified Multi-Task Metric Learning.}
We employ a ResNet-18~\cite{resnet18} $E_\theta$ as the visual encoder to map the input observation image $I$ into a feature representation $f(I) \in \mathbb{R}^d$. To enforce discriminative feature boundaries, we optimize $E_\theta$ via the Triplet Margin Loss~\cite{triletloss}:
\begin{equation}
\mathcal{L}_{\text{triplet}} = \max(0, d(I_a, I_p) - d(I_a, I_n) + m).
\end{equation}
Specifically, we select a ground-truth keyframe as the anchor $I_a$, characterized by the tuple $(i, \psi, e)$ denoting task, phase, and episode respectively. The \textbf{positive} $I_p$ is sampled from a different episode of the same phase $(i, \psi, e')$ where $e' \neq e$. This forces the encoder to learn an invariant representation for each phase across different episodes. For the \textbf{negative} $I_n$, we employ a balanced sampling strategy where three categories of negatives are drawn with equal probability:
\begin{itemize}[leftmargin=*, topsep=0pt, itemsep=2pt]
\item \textbf{Temporal Neighbors} $\mathcal{N}_{\text{adj}}$: Frames $I_{t'}$ within the same episode whose temporal offset from the anchor satisfies $\delta_{\min} \le |t' - t| \le \delta_{\max}$, enforcing fine-grained discrimination against visually similar neighbors.
\item \textbf{Intra-task Phase Negatives} $\mathcal{N}_{\text{phase}}$: Keyframes sampled from the same task $i$ but a different phase $\psi' \neq \psi$, ensuring the encoder captures the unique semantic progression within a task.
\item \textbf{Inter-task Negatives} $\mathcal{N}_{\text{task}}$: Keyframes sampled from a different task $j \neq i$, ensuring global task discriminability.
\end{itemize}
This tri-modal negative sampling forces the latent space to simultaneously respect temporal sequence, phase identity, and task boundaries.

\noindent\textbf{Stage II: Task-Modulated Query Matching Training.} During inference, we freeze the visual encoder $E_\theta$ trained in Stage I and employ a lightweight query network to detect semantic keyframes. First, to incorporate local temporal context, we aggregate a sliding window of features into a cohesive representation $\mathbf{H}_{\text{vis}}$:\begin{equation}\mathbf{H}_{\text{vis}} = \text{SelfAttn}\left( \text{Concat}[E_\theta(\mathbf{I}_{t-k:t})] \right).\end{equation}
Next, we synthesize a dynamic logic query $\mathbf{q}_{\text{logic}}$. Specifically, we employ two sets of learnable embeddings to represent task identities and generic phase concepts. Given the natural language instruction $l$, we map it to the corresponding task identity to select the specific embedding $e_{\text{task}}$. We then project the task-ID embedding $e_{\text{task}}$ into affine parameters $[\gamma, \beta]$ via a generator $g_\phi$:
\begin{equation}
    [\gamma, \beta] = g_\phi(e_{\text{task}}).
\end{equation}
These parameters then modulate the shared phase embedding $e_{\text{phase}}$ via FiLM~\cite{film} to encode task-specific semantics:
\begin{equation}
    \mathbf{q}_{\text{logic}} = \gamma \odot e_{\text{phase}} + \beta.
\end{equation}
Finally, the query $\mathbf{q}_{\text{logic}}$ retrieves relevant cues from $\mathbf{H}_{\text{vis}}$ via Cross-Attention. We compute a matching score $s_t$:
\begin{equation}
    s_t = \text{MLP}(\text{CrossAttn}(Q=\mathbf{q}_{\text{logic}}, K{=}V{=}\mathbf{H}_{\text{vis}})).
\end{equation}
When the matching score $s_t$ exceeds the pre-defined confidence threshold $\tau_{\text{conf}}$, the system identifies a milestone completion. Consequently, the current observation $o_t$ is retrieved and appended to the sparse history buffer, and the phase pointer is updated ($id_{\text{phase}} \leftarrow id_{\text{phase}} + 1$).

\noindent\textbf{Balanced Training Protocol.} To mitigate keyframe sparsity and enhance robustness, we implement a phase-guided balanced sampling strategy to generate training pairs with binary target labels $y \in \{0, 1\}$:
\begin{itemize}[leftmargin=*, noitemsep, topsep=0pt, parsep=5pt, partopsep=0pt]
\item \textbf{Positives ($y=1.0$):} Frames sampled at the ground-truth keyframe and its immediate temporal neighbors (offsets $\delta \in \{-1, 0, 1\}$) to accommodate minor annotation jitter.
\item \textbf{In-trajectory Negatives ($y=0.0$):} Frames randomly sampled from $M$ equidistant intervals within the trajectory segment $[k_{i-1}, k_i]$ to suppress false triggers.
\item \textbf{Phase-mismatched Negatives ($y=0.0$):} Pairs combining a valid milestone image $o_{k_i}$ with a query $\mathbf{q}(id_{phase}+1)$ from the subsequent phase, enforcing strict sequential logic.
\end{itemize}

\noindent\textbf{Inference and Greedy Smoothing.} To ensure robust milestone detection against score fluctuations during real-time execution, we implement a greedy temporal smoothing mechanism with a reset capability. When the matching score satisfies $s_t > \tau_{\text{conf}}$, the current observation is cached as a provisional keyframe, triggering the entry into a temporal validation window. This window functions as a stability test: if any subsequent frame within the window again exceeds $\tau_{\text{conf}}$, it acts as a stronger candidate, immediately superseding the current provisional keyframe and resetting the validation timer. This mechanism ensures the system greedily latches onto the latest valid frame to capture the exact state of subtask completion. The candidate is committed to the finalized history buffer $\mathcal{O}_{\text{key}}$ only after the score consistently remains below $\tau_{\text{conf}}$ for the full duration of the window, at which point the phase pointer increments.

\subsection{Keyframe-Chaining Policy Architecture}

We adopt the state-of-the-art generalist policy GR00T-N1.5 as our backbone to validate the efficacy of the proposed framework. Its transformer-based architecture intrinsically supports variable-length token sequences, making it an ideal substrate for integrating our non-uniform sparse history.
To effectively leverage the extracted keyframes, we reformulate the perception stream of GR00T-N1.5 by transitioning from redundant dense observations to a Sparse Semantic History. Specifically, we construct a non-uniform input sequence $\mathcal{I} = \{ o_{k_1}, \dots, o_{k_n}, o_t \}$, concatenating retrieved historical keyframes with the current visual observation. To facilitate precise reasoning over this temporal sequence, we employ a structured textual system prompt that explicitly underscores temporal relations and guides the model to analyze task progression. Finally, this semantically aligned sequence $\mathcal{I}$ is injected as a condition into the Flow Matching action head~\cite{flowmatching}. By grounding the policy in the task's topological progress, this architecture resolves the state aliasing inherent in single-frame observations, empowering the agent to transcend Markovian limitations and generate actions coherent with the complete logical chain.

\section{Experiments}
In this section, we conduct a series of comprehensive experiments across high-fidelity simulation and real-world environments to systematically evaluate the efficacy of our Keyframe-Chaining VLA framework. The experimental design focuses on the following three dimensions:

\begin{enumerate}[topsep=0pt, partopsep=0pt, itemsep=2pt, parsep=0pt] \item \textbf{Effectiveness of Sparse Semantic History:} Can our proposed sparse history mechanism effectively mitigate state aliasing in non-Markovian tasks, outperforming baselines that rely on dense observation windows or naive sampling heuristics? \item \textbf{Accuracy and Necessity of KSM:} What is the diagnostic precision of the Keyframe Selection Module (KSM) in detecting semantic milestones, and are its internal components essential for establishing a discriminative feature space? \item \textbf{Physical Deployment and Robustness:} For policies trained directly in the real world, to what extent can the sparse keyframe-chaining mechanism maintain stable long-horizon control under inherent environmental noise and perceptual disturbances? \end{enumerate}

\subsection{Simulation: Memory Dependence Benchmark}
\label{sec:tasks} 

\begin{figure*}[t]
    \centering
    \includegraphics[width=0.95\textwidth]{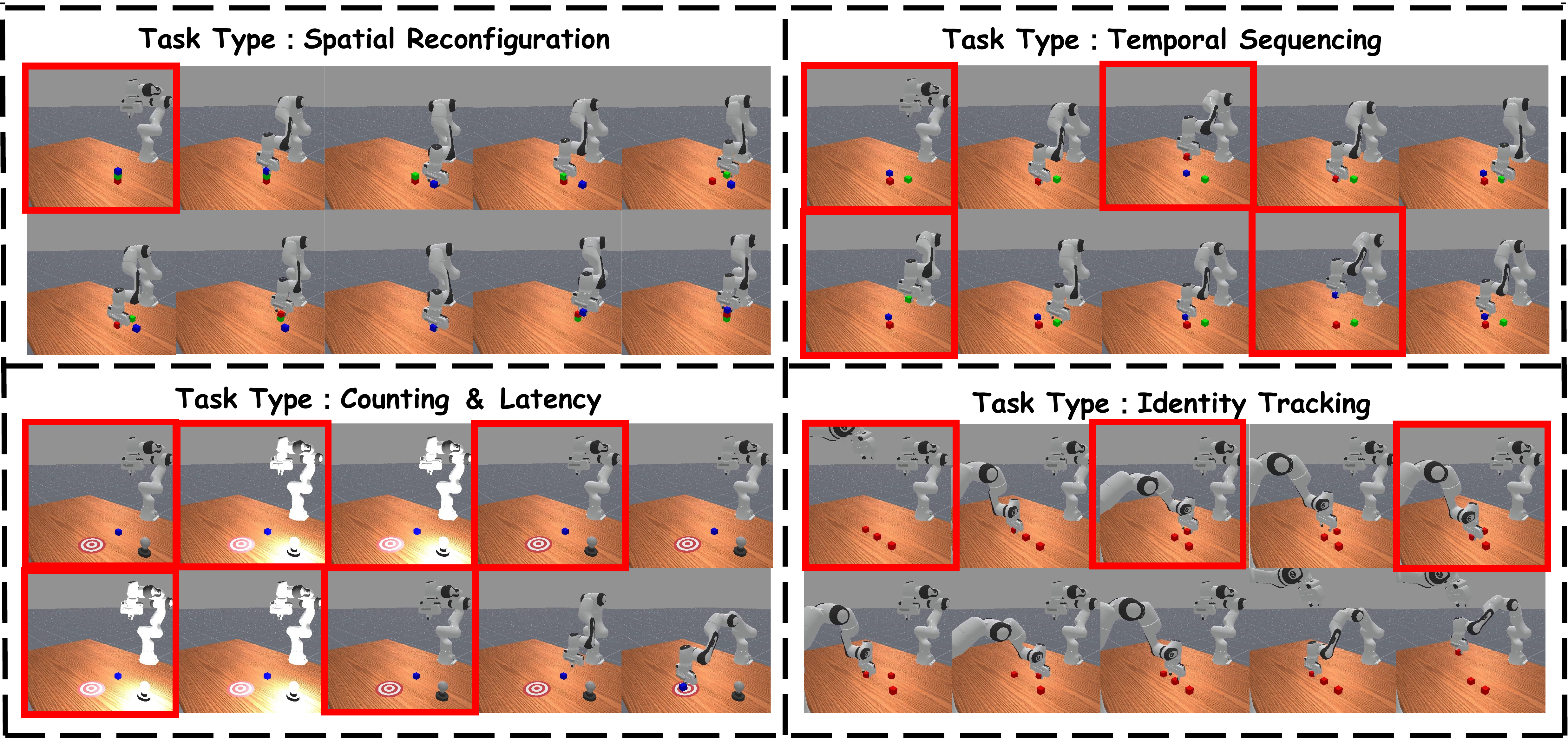}
\caption{
\textbf{Overview of the four custom Non-Markovian ManiSkill tasks.}
Red bounding boxes indicate the ground-truth keyframes utilized for training. Please refer to the \textbf{Appendix} for the detailed definitions of keyframe events for each task.
}
    \label{fig:maniskill}
\end{figure*}

\begin{table*}[t]
\footnotesize
\centering
\caption{
Quantitative comparison of success rates (\%) on non-Markovian tasks.
Keyframe-Chaining VLA(ours) achieves robust and superior performance.
}
\resizebox{\textwidth}{!}{%
\begin{tabular}{@{}llccccccc@{}}
\toprule
\textbf{Model / Configuration} & \textbf{Sampling} & $\mathbf{N_h}$ & $\mathbf{I}$ & \textbf{Spatial} & \textbf{Temporal} & \textbf{Identity} & \textbf{Counting} & \textbf{Average} \\ \midrule

$\pi_0$ ~\cite{pi0} & \multirow{3}{*}{Dense} & 0 & - & 2.0 & 0.0 & 0.0 & 60.0 & 15.5 \\ 
Diffusion Policy ~\cite{dp} & & 1 & 1 & 22.0 & 10.0 & 0.0 & 30.0 & 15.5 \\ 
GR00T-N1.5~\cite{gr00t} (\textit{No History}) & & 0 & - & 20.0 & 0.0 & 28.0 & 16.0 & 16.0 \\ 
\midrule
\multirow{3}{*}{GR00T-N1.5 (\textit{Short-term})} & \multirow{3}{*}{Dense} & 1 & 1 & 8.0 & 16.0 & 30.0 & 4.0 & 14.5 \\
 & & 2 & 1 & 18.0 & 52.0 & 12.0 & 0.0 & 20.0 \\
 & & 3 & 1 & 14.0 & 54.0 & 0.0 & 40.0 & 27.0 \\ 
\midrule

\multirow{11}{*}{GR00T-N1.5 (\textit{Long-term})} & \multirow{11}{*}{Fixed Stride} & 3 & 5  & 20.0 & 80.0 & 32.0 & 30.0 & 40.5 \\
 & & 3 & 10 & 18.0 & \underline{94.0} & 26.0 & 60.0 & 49.5 \\
 & & 3 & 20 & 18.0 & \underline{94.0} & 22.0 & \underline{74.0} & 52.0 \\
 & & 3 & 30 & 6.0  & 38.0 & 44.0 & 14.0 & 25.5 \\
 & & 3 & 40 & \underline{60.0} & 84.0 & \underline{84.0} & 0.0  & \underline{57.0} \\
 & & 3 & 50 & 16.0 & 66.0 & 52.0 & 60.0 & 48.5 \\
 & & 3 & 60 & 26.0 & 70.0 & 78.0 & 2.0  & 44.0 \\
 & & 3 & 70 & 24.0 & 40.0 & 32.0 & 2.0  & 25.0 \\
 & & 3 & 80 & 0.0 & 22.0 & 8.0 & 22.0 & 13.0 \\
 & & 3 & 90 & 28.0 & 52.0 & 50.0 & 8.0 & 35.0 \\
 & & 3 & 100 & 2.0 & 14.0 & 4.0 & 4.0 & 6.0 \\ 
\midrule
\textbf{Keyframe-Chaining VLA (ours)} & \textbf{Keyframes} & - & - & \textbf{70.0} & \textbf{98.0} & \textbf{100.0} & \textbf{100.0} & \textbf{92.0} \\ \bottomrule
\end{tabular}%
}
\vspace{2pt}
\label{tab:main_results}
\end{table*}

\noindent \textbf{Bridging the Simulation Evaluation Gap.} 
A critical bottleneck in advancing memory-based robot learning is the absence of suitable evaluation platforms. Existing benchmarks, such as LIBERO~\cite{libero}, RLBench~\cite{rlbench}, and CALVIN~\cite{calvin}, typically focus on procedural complexity but remain predominantly \textit{Markovian}—where the optimal action can be inferred solely from the current observation (e.g., ``open the visible drawer"). This ``Markovian bias" in current datasets fails to penalize myopic policies, masking their inability to handle state aliasing.

To address this deficiency, we developed and open-sourced a specialized \textbf{Long-Horizon Memory dependence Benchmark} built upon the high-fidelity ManiSkill~\cite{maniskill2} simulator. Detailed specifications regarding environment initialization, success metrics, and ground-truth keyframe definitions are provided in Appendix~\ref{sec:task_details}. Unlike previous suites, our benchmark introduces \textit{Hard Non-Markovian Constraints}, where critical task information (e.g., past interaction history) is strictly hidden from the current observation. This suite serves as a rigorous testbed for the community to evaluate agents' capacity for maintaining persistent internal states and temporal reasoning.

\noindent \textbf{Task Protocols.} We designed four tasks (visualized in Fig.~\ref{fig:maniskill}) that necessitate distinct types of memory dependencies:

\begin{itemize}[topsep=0pt, partopsep=0pt, itemsep=4pt, parsep=0pt]
    \item \textbf{Spatial Reconfiguration}: The agent must dismantle a vertical stack of three randomly ordered blocks and reconstruct them in a permuted sequence. \textbf{Rationale}: The primary difficulty lies in the ``destruction-reconstruction'' cycle. Once disassembled, visual information regarding the initial relative order is permanently lost. The agent must rely entirely on its memory of the initial configuration frame to execute the correct re-stacking sequence.

    \item \textbf{Temporal Sequencing}: The robot must perform a ``pick-lift-reset'' cycle for three colored cubes strictly in the order of $Red \rightarrow Green \rightarrow Blue$. \textbf{Rationale}: This task introduces severe state aliasing regarding task progression. For instance, a hovering gripper is insufficient to determine which blocks have already been manipulated from a static observation. The agent must verify the historical completion of subgoals via retrieved keyframes to distinguish valid progression from sequence violations.

    \item \textbf{Counting \& Latency}: A signal lamp flashes twice with a randomized interval. The agent must count these pulses and push the target only after the second flash. \textbf{Rationale}: This evaluates the capacity to maintain internal states during static intervals. Since the scene during the inter-flash gap is indistinguishable from the initial state, a Markovian policy cannot determine the current count without persistent historical context.

    \item \textbf{Identity Tracking}: Three visually identical red blocks are aligned, and an auxiliary arm performs a rapid swap between two of them. The agent is tasked with picking the specific block that was originally in the center. \textbf{Rationale}: As the objects share identical visual features, the target's identity is defined solely by its history. The model must perform object constancy reasoning by retrieving the specific historical frame capturing the swap event, a capability fundamentally unattainable without precise temporal recall.
\end{itemize}

\textbf{Baselines.} We compare our Keyframe-Chaining VLA against a representative subset of state-of-the-art robot policies, evaluating them under various temporal configurations to establish a robust benchmark:
\begin{itemize}[leftmargin=*, topsep=2pt, itemsep=2pt, parsep=0pt]
    \item \textbf{Diffusion Policy (DP)} \cite{dp}: A leading behavior cloning approach that models the action distribution using conditional diffusion processes. We utilize the standard implementation with a short observation horizon ($T_{obs}=2$) to represent strong short-term policies.

    \item \textbf{$\pi_0$ (Flow Matching)} \cite{pi0}: A state-of-the-art generalist VLA policy that employs flow matching for high-fidelity action generation. Similar to DP, it primarily relies on immediate visual context for decision-making.

    \item \textbf{GR00T-N1.5} \cite{gr00t}: A large-scale generalist VLA. To systematically evaluate the impact of temporal receptive fields, we conduct a comprehensive analysis of GR00T under three distinct memory regimes defined by history length ($N_h$) and sampling interval ($I$):
    (i) No History ($N_h=0$): A pure Markovian baseline relying solely on the current frame.
    (ii) Short-term History ($N_h \in \{1, \dots, 3\}, I=1$): A dense observation window capturing local temporal cues.
    (iii) Long-term History ($N_h=3, I \in \{5, \dots, 100\}$): A fixed-stride sampling approach designed to extend the horizon coverage at the cost of temporal resolution.
\end{itemize}

\textbf{Quantitative Analysis.} 
As presented in Table~\ref{tab:main_results}, Keyframe-Chaining VLA demonstrates superior performance with a mean success rate of \textbf{92.0\%}, surpassing the strongest baseline (57.0\%) by a significant margin. A closer examination of baseline behaviors reveals the fundamental limitations of standard memory regimes. Specifically, policies relying on short, dense observation windows ($N_h \le 3, I=1$) consistently fail across all non-Markovian tasks, averaging below 30\% success. This widespread failure underscores that simply stacking immediate observations is insufficient for long-horizon reasoning, as task-critical cues—such as the initial block configuration in \textit{Spatial Reconfiguration}—often recede far beyond this narrow temporal horizon.

Furthermore, while extending the receptive field via fixed-stride sampling ($I > 1$) attempts to mitigate this myopia, it introduces an inherent trade-off between \textbf{temporal horizon} and \textbf{resolution}. This conflict is clearly evidenced by the performance divergence across tasks: a coarse sampling rate (e.g., $I=40$) successfully captures the distant initial state for \textit{Spatial Reconfiguration} (60\% success) but fails completely in \textit{Counting} (0\%) due to signal aliasing, where rapid light flickers are skipped between frames. Conversely, a finer stride ($I=20$) resolves the high-frequency counting signals (74\%) but fails to span the necessary history for spatial reasoning (18\%). Consequently, no single fixed sampling rate can generalize across tasks with distinct temporal dynamics.

In contrast, our method circumvents this hyperparameter trade-off by shifting from fixed-interval scanning to event-driven retrieval. By explicitly selecting semantically significant frames, the model achieves consistent high performance (e.g., 100\% in Identity and Counting) regardless of the event's duration or timing. This result suggests that the bottleneck in long-horizon manipulation is not merely the capacity of memory, but the semantic relevance of the retrieved context, validating our hypothesis that sparse, discriminative history is a more effective representation than dense, noisy windows.

\textbf{Implementation Details.} 
We curated a multi-task dataset comprising 400 episodes, balanced with 100 episodes for each non-Markovian task. The policy architecture integrates a 3B-parameter VLM backbone for semantic reasoning with a 0.5B-parameter DiT head for precise action generation. Training was conducted on a single NVIDIA L40 GPU for 50,000 steps with a batch size of 16, utilizing the AdamW optimizer with a constant learning rate of $1 \times 10^{-4}$.

\subsection{KSM Evaluation and Ablations}
\label{sec:ksm_ablation}

\noindent \textbf{Diagnostic Setup.} 
We curated a centralized dataset of 400 episodes (100 per task) to evaluate the Keyframe Selection Module (KSM). An 80/20 split partitions the data into training sets for the KSM task-modulated query network and held-out test sets to verify generalization to unseen execution trajectories.

\noindent \textbf{Metrics Definition.} 
Applying the temporal greedy smoothing mechanism, we cluster raw trigger indices within a 5-frame window and utilize the cluster median as the representative timestamp. A prediction is classified as a True Positive (TP) if the temporal offset between the cluster median and the Ground Truth (GT) lies within a tolerance of $\pm 10$ frames. False Positives (FP) indicate redundant or misaligned triggers, while False Negatives (FN) denote missed keyframes. 
Based on these classifications, we report Precision, Recall, and the F1-score. Additionally, we analyze error modes using the False Negative Rate and False Positive Rate.

\vspace{-10pt}
\begin{table}[ht]
\centering
\caption{
Comparative diagnostic evaluation of KSM paradigms.
Metrics are averaged across four tasks and reported in percentage. All metrics are reported in percentage (\%).
Detailed task-wise results are provided in Appendix~\ref{sec:task_wise_ksm}.
}
\resizebox{\columnwidth}{!}{%
\begin{tabular}{@{}lccccc@{}}
\toprule
\textbf{KSM Training Paradigm} & \textbf{P} (\%) $\uparrow$ & \textbf{R} (\%) $\uparrow$ & \textbf{F1} (\%) $\uparrow$ & \textbf{FPR} (\%) $\downarrow$ & \textbf{FNR} (\%) $\downarrow$ \\ \midrule
w/o Metric Pre-training    & 90.7 & 81.7 & 85.1 & 9.3 & 18.3 \\
Joint End-to-End           & 93.6 & 79.6 & 83.4 & 6.4 & 20.4 \\
\textbf{Ours (Two-stage)}  & \textbf{97.5} & \textbf{97.5} & \textbf{97.5} & \textbf{2.5} & \textbf{2.5} \\ \bottomrule
\end{tabular}%
}
\label{tab:ksm_diagnostic}
\end{table}

\noindent \textbf{Analysis of Training Paradigms.} 
Table \ref{tab:ksm_diagnostic} highlights the efficacy of our decoupled training strategy, which achieves a near-perfect F1-score of 97.5. 
Crucially, the Joint End-to-End paradigm exhibits a high False Negative Rate (20.4). This degradation is most pronounced in the \textit{Counting} task (Recall=40.0), indicating that without the discriminative embedding space enforced by the triplet loss, the model fails to disentangle transient semantic events from the static background. 
By strictly enforcing metric learning before query training, our Two-stage approach ensures the VLA receives a high-fidelity sparse history $\mathcal{O}_{\text{key}}$.

\vspace{-10pt}
\begin{table}[ht]
\centering
\caption{
\textbf{Ablation on Prompt Refinement.} 
Comparing context-aware prompting against generic instructions. 
}
\resizebox{\columnwidth}{!}{%
\begin{tabular}{@{}lccccc@{}}
\toprule
\textbf{Configuration} & \textbf{Spatial} & \textbf{Temporal} & \textbf{Identity} & \textbf{Counting} & \textbf{Average} \\ \midrule
w/o Prompt Refinement & 56.0 & 96.0 & 100.0 & 100.0 & 88.0 \\
\textbf{Ours (Context-Aware)} & \textbf{70.0} & \textbf{98.0} & \textbf{100.0} & \textbf{100.0} & \textbf{92.0} \\ \bottomrule
\end{tabular}%
}
\label{tab:prompt_ablation}
\end{table}

\noindent \textbf{Necessity of Prompt Refinement.} 
We isolate the impact of context-aware prompting by comparing our refined prompts against generic instructions. 
Table \ref{tab:prompt_ablation} reveals a task-dependent sensitivity: \textit{Spatial Reconfiguration} suffers a significant drop (70\% $\rightarrow$ 56\%) without refinement. 
To achieve this, we design a structured system prompt that explicitly defines the input as a chronological sequence of multi-view image pairs. The exact prompt formulation is provided in Appendix~\ref{sec:appendix_prompts}.

\subsection{Real-World Experiments}
\label{sec:real_world}

To validate the effectiveness of our approach in practical settings, we conduct real-world experiments using an AgileX Piper robotic arm. We focus on evaluating whether the Keyframe-Chaining VLA can sustain long-horizon reasoning in physical environments. For policy adaptation, we collected \textbf{50 expert demonstrations for each task} via teleoperation using a puppet-arm setup. Visualizations of the execution rollouts for these tasks are provided in Appendix~\ref{sec:appendix_real_world_viz}.

\begin{figure}[ht]
    \centering
    \includegraphics[width=1.0\columnwidth]{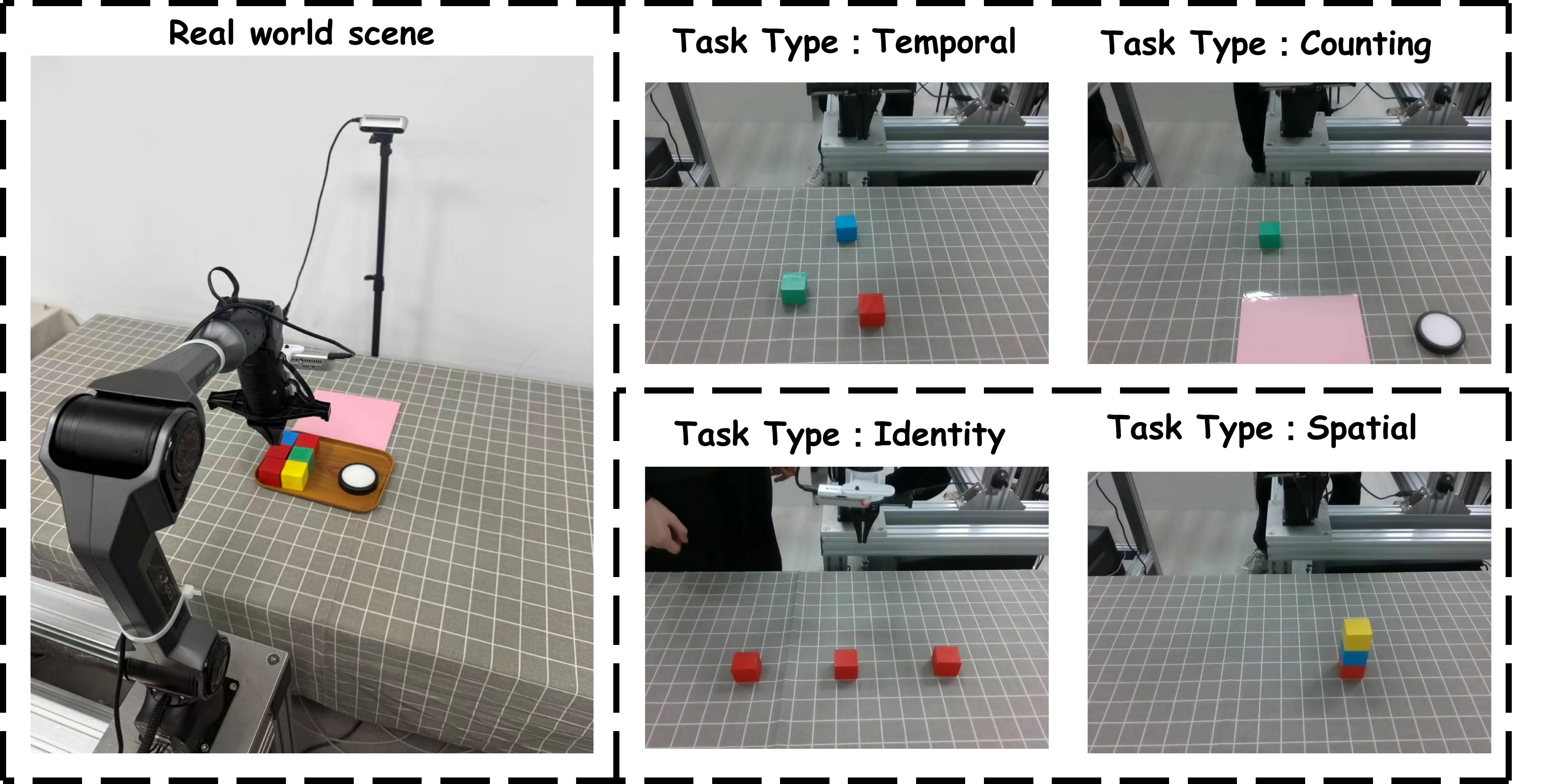} 
    \caption{
Real-world Experimental Setup.
    }
    \label{fig:real_world_setup}
\end{figure}

\vspace{-10pt}

\begin{table}[ht]
\centering
\renewcommand{\arraystretch}{1.25} 
\caption{
\textbf{Real-world quantitative evaluation (20 trials per task).} 
Results are reported as \textbf{Success Rate} / {Completion Rate} (\%). 
\textit{Completion Rate} measures the percentage of completed sub-stages (see Appendix for stage definitions). 
Our method demonstrates decisive advantages in multi-stage reasoning tasks.
}

\setlength{\tabcolsep}{2.5pt}
\resizebox{\columnwidth}{!}{%
\begin{tabular}{@{}lccccc@{}}
\toprule
\textbf{Model} & \textbf{Spatial} & \textbf{Temporal} & \textbf{Counting} & \textbf{Identity} & \textbf{Average} \\ 
& \scriptsize{\textit{(4 stages)}} & \scriptsize{\textit{(3 stages)}} & \scriptsize{\textit{(2 stages)}} & \scriptsize{\textit{(2 stages)}} & \\
\midrule

DP & 0.0 / 33.8 & 0.0 / 16.7 & 5.0 / 35.0 & 10.0 / 45.0 & 3.8 / 32.6 \\
GR00T & 0.0 / 47.5 & 0.0 / 30.0 & 10.0 / 45.0 & 15.0 / 50.0 & 6.3 / 43.1 \\

\rowcolor{mygray} 
\textbf{Ours} & \textbf{10.0 / 58.8} & \textbf{50.0 / 75.0} & \textbf{80.0 / 90.0} & \textbf{55.0 / 77.5} & \textbf{48.75 / 75.3} \\ 
\bottomrule
\end{tabular}%
}
\label{tab:real_world_results}
\end{table}

\noindent \textbf{Quantitative Analysis.} Evaluated over 20 trials, our method establishes a decisive performance margin. 
Across the evaluated tasks, our approach achieves consistent improvements in both Success Rate (SR) and Completion Rate (CR). Unlike baselines that frequently succumb to early-stage errors, our policy maintains semantic consistency over long horizons, securing significantly higher SR and demonstrating superior stage-wise progress even in partial completions.
Overall, this comprehensive superiority validates the framework's improved geometric and temporal grounding in physical environments.

\section{Conclusion}

We presented \textbf{Keyframe-Chaining VLA}, resolving non-Markovian ambiguity via Sparse Semantic History. Our Task-Modulated KSM extracts event-driven keyframes to efficiently ground long-horizon dependencies, achieving a 92.0\% success rate on our ManiSkill Benchmark and robust real-world performance.
\textbf{Limitations and Future Work.} Storing raw visual keyframes incurs pixel-space redundancy and scales linearly. Future work will investigate compressing history into compact latent abstractions or discrete semantic tokens to distill task-relevant signals, alongside exploring dynamic memory update mechanisms.

\nocite{langley00}

\bibliography{references}
\bibliographystyle{icml2025}

\newpage
\appendix
\onecolumn
\section{Task Specifications and Oracle Design}
\label{sec:task_details}

In this section, we detail the implementation logic, success criteria, and ground-truth (oracle) keyframe selection heuristics for the four custom non-Markovian tasks designed in ManiSkill.

\subsection{Temporal Sequencing (PickPlaceThreeTimes)}
\textbf{Instruction:} ``First, pick up the red cube and place it back on the table. Next, do the same for the green cube. Finally, the blue cube.''

\begin{itemize}[leftmargin=*]
    \item \textbf{Environment Setup:} Three colored cubes (Red, Green, Blue) are initialized at random poses on the tabletop. The target positions for placement coincide with their initial coordinates.
    \item \textbf{Success Criteria:} The task is considered successful if and only if:
    \begin{enumerate}
        \item \textit{Sequential Lifting:} The Red, Green, and Blue cubes are detected to exceed a height threshold $h_{lift}$ strictly in the order of $Red \rightarrow Green \rightarrow Blue$.
        \item \textit{Precise Placement:} Upon task completion, the Euclidean distance between each cube's final position and its initial position is within a tolerance threshold $\epsilon$ (i.e., returned to origin).
    \end{enumerate}
    \item \textbf{Oracle Keyframe Definition (4 Frames):}
    To train the KSM, we select the following ground-truth keyframes based on the z-axis trajectory of the objects:
    \begin{enumerate}
        \item \textbf{Frame 0:} The initial observation containing the goal configuration.
        \item \textbf{Frame $t_{R}$:} The timestamp when the \textit{Red} cube reaches its maximum height ($z_{max}$).
        \item \textbf{Frame $t_{G}$:} The timestamp when the \textit{Green} cube reaches its maximum height.
        \item \textbf{Frame $t_{B}$:} The timestamp when the \textit{Blue} cube reaches its maximum height.
    \end{enumerate}
\end{itemize}

\subsection{Counting \& Latency (PushCubeWithSignal)}
\textbf{Instruction:} ``Wait for the signal light to flash twice, then push the cube to the target.''

\begin{itemize}[leftmargin=*]
    \item \textbf{Environment Setup:} A cube is initialized at a random position. A signal lamp undergoes a state transition sequence: $Off \rightarrow On \rightarrow Off \rightarrow On \rightarrow Off$. The duration of the first four states is randomized to prevent temporal overfitting.
    \item \textbf{Success Criteria:} 
    \begin{enumerate}
        \item \textit{Goal Achievement:} The cube is pushed to the target region.
        \item \textit{Constraint Satisfaction:} The cube's displacement must remain zero until the signal light enters the final $Off$ state (i.e., after the second flash completes). Any movement prior to this trigger results in immediate failure.
    \end{enumerate}
    \item \textbf{Oracle Keyframe Definition (5 Frames):}
    Keyframes are selected at the exact moments of signal state transitions (Signal Edges):
    \begin{enumerate}
        \item \textbf{Frame 0:} Initial state.
        \item \textbf{Frame $t_{1}$:} Edge rising (1st flash starts: $Off \rightarrow On$).
        \item \textbf{Frame $t_{2}$:} Edge falling (1st flash ends: $On \rightarrow Off$).
        \item \textbf{Frame $t_{3}$:} Edge rising (2nd flash starts: $Off \rightarrow On$).
        \item \textbf{Frame $t_{4}$:} Edge falling (2nd flash ends: $On \rightarrow Off$).
    \end{enumerate}
\end{itemize}

\subsection{Spatial Reconfiguration (SwapPosition)}
\textbf{Instruction:} ``Swap the position of the bottom and middle cubes.''

\begin{itemize}[leftmargin=*]
    \item \textbf{Environment Setup:} Three colored cubes (Red, Green, Blue) are initialized in a single vertical stack. The stacking order and base position are randomized.
    \item \textbf{Success Criteria:} The task checks the relative vertical ordering (z-coordinates) of the cubes:
    \begin{enumerate}
        \item The cube originally at the \textit{Top} must remain at the highest z-position.
        \item The cube originally at the \textit{Bottom} must be at the middle z-position.
        \item The cube originally in the \textit{Middle} must be at the lowest z-position.
    \end{enumerate}
    \item \textbf{Oracle Keyframe Definition (1 Frame):}
    Keyframes capture the critical disassembly steps required to memorize the initial configuration:
    \begin{enumerate}
        \item \textbf{Frame 0:} The initial stacked state (crucial for identifying the original order).
    \end{enumerate}
\end{itemize}

\subsection{Identity Tracking (TeacherArmShuffle)}
\textbf{Instruction:} ``After the cubes are swapped, pick up the cube that was originally in the middle.''

\begin{itemize}[leftmargin=*]
    \item \textbf{Environment Setup:} Three visually identical red cubes are aligned horizontally. An automated ``Teacher Arm" performs a random shuffle operation on two randomly selected cubes (denoted as Cube A and Cube B). The shuffle sequence is: (1) Move A to a temporary buffer zone. (2) Move B to A's original position. (3) Move A to B's original position.
    \item \textbf{Success Criteria:} After the Teacher Arm completes the shuffle sequence and retracts, the agent must grasp and lift the specific object ID that corresponded to the geometric center at $t=0$, exceeding a height threshold $h_{lift}$.
    \item \textbf{Oracle Keyframe Definition (3 Frames):}
    Keyframes focus on the swap trajectory to resolve object identity:
    \begin{enumerate}
        \item \textbf{Frame 0:} Initial state establishing the ``Middle" identity.
        \item \textbf{Frame $t_{A}$:} The moment the Teacher Arm places Cube A into the temporary buffer zone.
        \item \textbf{Frame $t_{B}$:} The moment the Teacher Arm grasps Cube B (signifying the start of the second displacement).
    \end{enumerate}
\end{itemize}

\section{Task-wise KSM Diagnostic Analysis}
\label{sec:task_wise_ksm}

In Section \ref{sec:ksm_ablation}, we reported the aggregated diagnostic metrics for the Keyframe Selection Module (KSM). Here, we provide the granular breakdown of performance across all four tasks, comparing our proposed Two-stage approach against the ``End-to-End" and ``w/o Metric Pre-training" baselines.

As shown in Table \ref{tab:appendix_ksm_breakdown}, the **Counting** task proves most challenging for baseline methods due to the transient nature of the light signal (rapid flickering). The \textit{End-to-End} model suffers a severe drop in Recall (40.0\%) on this task, confirming that without the explicit metric constraint from Stage I, the model struggles to distinguish the semantic ``signal" from background noise. In contrast, our method maintains near-perfect detection (100.0\%) across all tasks, except for a slight dip in \textit{Spatial Reconfiguration} (90.0\%), which involves complex occlusion during the destruction-reconstruction cycle.

\begin{table}[H] 
\footnotesize
\centering
\renewcommand{\arraystretch}{1.2}
\caption{
\textbf{Detailed Task-wise KSM Diagnostic Metrics (\%).} 
We compare Precision (P), Recall (R), and F1-score across three training paradigms. 
Our method consistently achieves optimal or near-optimal performance across all categories.
}
\label{tab:appendix_ksm_breakdown}
\begin{tabular}{@{}llccc@{}}
\toprule
\textbf{Task} & \textbf{Method} & \textbf{Precision} & \textbf{Recall} & \textbf{F1-Score} \\ \midrule

\multirow{3}{*}{\textbf{Temporal Sequencing}} 
& w/o Metric Pre-training & 80.3 & 81.7 & 81.0 \\
& Joint End-to-End & 93.3 & 93.3 & 93.3 \\
& \textbf{Ours (Two-stage)} & \textbf{100.0} & \textbf{100.0} & \textbf{100.0} \\ \midrule

\multirow{3}{*}{\textbf{Counting \& Latency}} 
& w/o Metric Pre-training & \textbf{100.0} & 62.5 & 76.9 \\
& Joint End-to-End & \textbf{100.0} & 40.0 & 57.1 \\
& \textbf{Ours (Two-stage)} & \textbf{100.0} & \textbf{100.0} & \textbf{100.0} \\ \midrule

\multirow{3}{*}{\textbf{Identity Tracking}} 
& w/o Metric Pre-training & 95.0 & 95.0 & 95.0 \\
& Joint End-to-End & \textbf{100.0} & \textbf{100.0} & \textbf{100.0} \\
& \textbf{Ours (Two-stage)} & \textbf{100.0} & \textbf{100.0} & \textbf{100.0} \\ \midrule

\multirow{3}{*}{\textbf{Spatial Reconfiguration}} 
& w/o Metric Pre-training & 87.5 & 87.5 & 87.5 \\
& Joint End-to-End & 81.0 & 85.0 & 82.9 \\
& \textbf{Ours (Two-stage)} & \textbf{90.0} & \textbf{90.0} & \textbf{90.0} \\ \bottomrule

\end{tabular}%
\end{table}

\section{VLA System Prompts Details}
\label{sec:appendix_prompts}

To effectively ground the Keyframe-Chaining VLA in the temporal context, we construct a structured system prompt that guides the model to interpret the interleaved image sequence. 

Since our input stream consists of both historical keyframes and the current observation, and each timestamp contains multi-view data (Third-Person and Wrist views), the prompt must explicitly define this topology to prevent ambiguity. The full system prompt used in our experiments is presented in Box~\ref{box:system_prompt}.

\begin{figure}[ht]
\centering
\begin{tcolorbox}[colback=gray!10!white, colframe=gray!50!black, title=\textbf{System Prompt Template}]
\ttfamily
\small
Task: \textbf{\{Language Instruction\}}

\vspace{0.5em}
Note: The provided images are a chronological sequence of \textbf{image pairs}. 
For each time step, two images are provided: the first is the \textbf{Third-Person View}, and the second is the corresponding \textbf{Wrist View}.

The final pair represents the current state, while all preceding pairs are history.

Based on this multi-view sequence, analyze the robot's progress and determine the immediate next action.
\end{tcolorbox}
\caption{The structured system prompt used to condition the VLA policy. The \texttt{\{Language Instruction\}} placeholder is replaced by the specific natural language command for the current episode (e.g., ``Stack the cubes based on their initial vertical order").}
\label{box:system_prompt}
\end{figure}

\section{Qualitative Real-World Visualizations}
\label{sec:appendix_real_world_viz}

\begin{figure*}[ht]
    \centering
    \includegraphics[width=\textwidth]{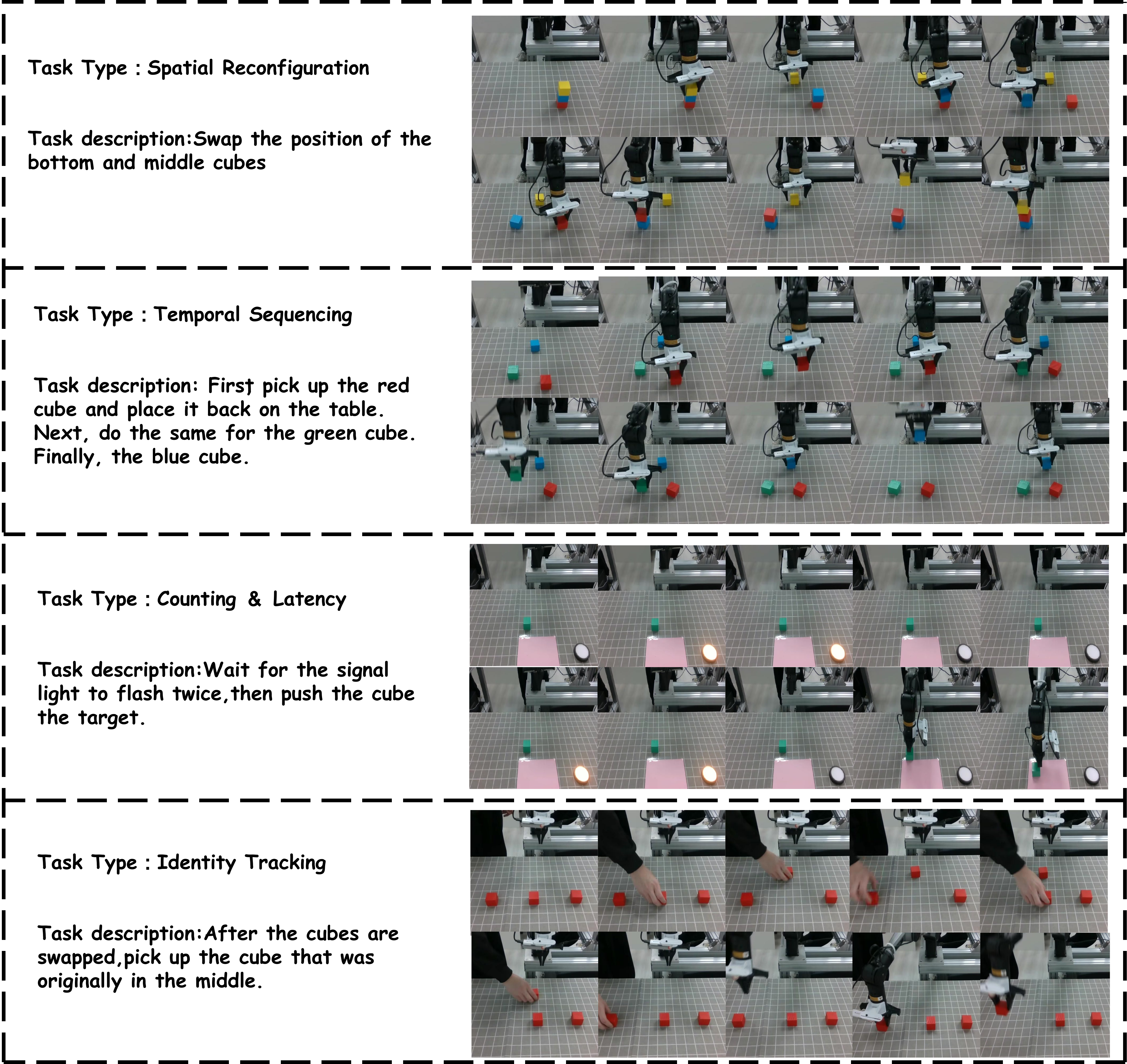} 
    \caption{
    \textbf{Qualitative rollouts of Keyframe-Chaining VLA in real-world scenarios.}
    Each row corresponds to one of the four Non-Markovian tasks.
    The snapshots demonstrate the agent's ability to maintain long-horizon consistency, such as waiting for signal lights (row 3) or tracking swapped objects (row 4).
    }
    \label{fig:real_world_rollouts}
\end{figure*}

To provide a concrete understanding of the system's behavior in physical environments, we present frame-by-frame execution sequences for all four Non-Markovian tasks in Figure~\ref{fig:real_world_rollouts}.

These visualizations illustrate how the agent handles distinct memory-dependent challenges:
\begin{itemize}
    \item \textbf{Spatial Reconfiguration:} The agent successfully restores the initial block permutation after the stack is destroyed.
    \item \textbf{Temporal Sequencing:} The policy strictly adheres to the $Red \rightarrow Green \rightarrow Blue$ sorting order constrained by the instruction.
    \item \textbf{Counting \& Latency:} The robot pauses correctly during the signal interval and executes the push only after the second flash.
    \item \textbf{Identity Tracking:} Despite the identical appearance of the blocks, the agent correctly identifies and retrieves the target block (originally in the center) after the human-actuated swap.
\end{itemize}

\clearpage
\section{Implementation Details and Hyperparameters}
\label{sec:appendix_hyperparams}

\begin{table}[h]
\centering
\footnotesize
\renewcommand{\arraystretch}{1.15} 

\begin{minipage}[t]{0.32\textwidth}
    \centering
    \caption{\textbf{VLA Fine-tuning}}
    \label{tab:vla_hyperparams}
    \resizebox{\linewidth}{!}{%
        \begin{tabular}{@{}ll@{}}
        \toprule
        \textbf{Parameter} & \textbf{Value} \\ \midrule
        Base Model & GR00T-N1.5-3B \\
        Optimizer & AdamW \\
        LR & $1 \times 10^{-4}$ \\
        Weight Decay & $1 \times 10^{-5}$ \\
        Batch Size & 16 (per GPU) \\
        Grad Accum & 1 \\
        Max Steps & 50,000 \\
        Warmup & 0.05 \\
        Scheduler & Cosine \\
        Backend & TorchCodec \\ \midrule
        \textit{Trainable} & \\
        \hspace{2mm}Projector & \checkmark \\
        \hspace{2mm}Diff. Head & \checkmark \\
        \hspace{2mm}Backbone & Frozen \\
        \bottomrule
        \end{tabular}%
    }
\end{minipage}
\hfill 
\begin{minipage}[t]{0.32\textwidth}
    \centering
    \caption{\textbf{KSM Stage I}}
    \label{tab:ksm_stage1_hyperparams}
    \resizebox{\linewidth}{!}{%
        \begin{tabular}{@{}ll@{}}
        \toprule
        \textbf{Parameter} & \textbf{Value} \\ \midrule
        Batch Size & 64 \\
        Epochs & 30 \\
        Optimizer & AdamW \\
        LR & $1 \times 10^{-4}$ \\
        Weight Decay & $1 \times 10^{-3}$ \\
        Loss & Triplet Margin \\
        Margin ($m$) & 1.0 \\
        Metric & Euclidean ($L_2$) \\
        \bottomrule
        \end{tabular}%
    }
\end{minipage}
\hfill 
\begin{minipage}[t]{0.32\textwidth}
    \centering
    \caption{\textbf{KSM Stage II}}
    \label{tab:ksm_stage2_hyperparams}
    \resizebox{\linewidth}{!}{%
        \begin{tabular}{@{}ll@{}}
        \toprule
        \textbf{Parameter} & \textbf{Value} \\ \midrule
        Context ($k$) & 3 frames \\
        Batch Size & 32 \\
        Epochs & 50 \\
        Optimizer & AdamW \\
        LR & $1 \times 10^{-4}$ \\
        Weight Decay & 0.05 \\
        Loss & BCEWithLogits \\
        Pos Weight & 5.0 \\
        \bottomrule
        \end{tabular}%
    }
\end{minipage}

\end{table}

\end{document}